\documentclass[11pt,a4paper]{article}
\usepackage[hyperref]{acl2021}
\usepackage{times}
\usepackage{amsfonts,amsmath,amssymb,amsthm}
\usepackage{latexsym}
\usepackage{booktabs}
\usepackage{graphicx}

\usepackage{tikz}
\usetikzlibrary{shapes,arrows,positioning,calc}

\usepackage{microtype}

\aclfinalcopy 

\title{Uncertainty and Surprisal Jointly Deliver the Punchline: Exploiting Incongruity-Based Features for Humor Recognition}

\author{Yubo Xie, Junze Li, and Pearl Pu \\
  School of Computer and Communication Sciences \\
  \'{E}cole Polytechnique F\'{e}d\'{e}rale de Lausanne \\
  Lausanne, Switzerland \\
  \texttt{\{yubo.xie,junze.li,pearl.pu\}@epfl.ch} \\}

\date{}

\begin{document}
\maketitle
\begin{abstract}
Humor recognition has been widely studied as a text classification problem using data-driven approaches. However, most existing work does not examine the actual joke mechanism to understand humor. We break down any joke into two distinct components: the set-up and the punchline, and further explore the special relationship between them. Inspired by the incongruity theory of humor, we model the set-up as the part developing semantic uncertainty, and the punchline disrupting audience expectations. With increasingly powerful language models, we were able to feed the set-up along with the punchline into the GPT-2 language model, and calculate the uncertainty and surprisal values of the jokes. By conducting experiments on the SemEval 2021 Task 7 dataset, we found that these two features have better capabilities of telling jokes from non-jokes, compared with existing baselines.
\end{abstract}

\section{Introduction}


One of the important aspects of computational humor is to develop computer programs capable of recognizing humor in text. Early work on humor recognition~\cite{DBLP:conf/naacl/MihalceaS05} proposed heuristic-based humor-specific stylistic features, for example alliteration, antonymy, and adult slang. More recent work~\cite{DBLP:conf/emnlp/YangLDH15,DBLP:conf/naacl/ChenS18,DBLP:conf/emnlp/WellerS19} regarded the problem as a text classification task, and adopted statistical machine learning methods and neural networks to train models on humor datasets. However, only few of the deep learning methods have tried to establish a connection between humor recognition and humor theories. Thus, one research direction in humor recognition is to bridge the disciplines of linguistics and artificial intelligence.

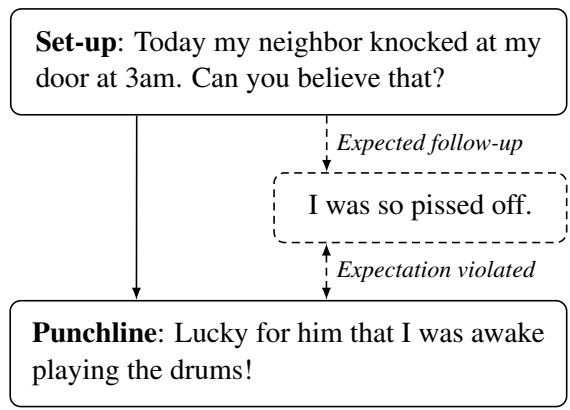
\begin{figure}[t]
    \centering
    \begin{tikzpicture}
        \node[rectangle,rounded corners,semithick,draw,align=left,inner sep=0.75em,minimum width=0.95\columnwidth] (setup) {\textbf{Set-up}: Today my neighbor knocked at my\\door at 3am. Can you believe that?};
        \node[rectangle,rounded corners,semithick,draw,densely dashed,align=left,inner sep=.75em,minimum width=0.5\columnwidth] (expect) [below=0.75cm of setup.south east,anchor=north east] {I was so pissed off.};
        \node[rectangle,rounded corners,semithick,draw,align=left,inner sep=0.75em,minimum width=0.95\columnwidth] (punch) [below=0.75cm of expect.south east,anchor=north east] {\textbf{Punchline}: Lucky for him that I was awake\\playing the drums!};
        \draw[-latex,semithick] ([xshift=-2cm]setup.south) -- ([xshift=-2cm]punch.north);
        \draw[-latex,densely dashed,semithick] ([xshift=0.5cm]setup.south) -- ++ (0,-0.75cm);
        \draw[latex-latex,densely dashed,semithick] ([xshift=0.5cm]punch.north) -- ++ (0,0.75cm);
        \node[below right=0.375cm and 0.5cm of setup.south,anchor=west] (a) {\textit{\small Expected follow-up}};
        \node[above right=0.375cm and 0.5cm of punch.north,anchor=west] (b) {\textit{\small Expectation violated}};
    \end{tikzpicture}
    \caption{A joke example consisting of a set-up and a punchline. A violation can be observed between the punchline and the expectation.}
    \label{fig:joke_example}
\end{figure}

In this paper, we restrict the subject of investigation to jokes, one of the most common humor types in text form.
As shown in Figure~\ref{fig:joke_example}, these jokes usually consist of a \emph{set-up} and a \emph{punchline}. The set-up creates a situation that introduces the hearer into the story framework, and the punchline concludes the joke in a succinct way, intended to make the hearer laugh. Perhaps the most suitable humor theory for explaining such humor phenomenon is the \emph{incongruity theory}, which states that the cause of laughter is the perception of something incongruous (the punchline) that violates the hearer's expectation (the set-up).

Based on the incongruity theory, we propose two features for humor recognition, by calculating the degree of incongruity between the set-up and the punchline. Recently popular pre-trained language models enable us to study such relationship based on large-scale corpora. Specifically, we fed the set-up along with the punchline into the GPT-2 language model~\cite{radford2019language}, and obtained the surprisal and uncertainty values of the joke, indicating how surprising it is for the model to generate the punchline, and the uncertainty while generating it. We conducted experiments on a manually labeled humor dataset, and the results showed that these two features could better distinguish jokes from non-jokes, compared with existing baselines. Our work made an attempt to bridge humor theories and humor recognition by applying large-scale pre-trained language models, and we hope it could inspire future research in computational humor.

\section{Related Work}

\paragraph{Humor Data}
\citet{DBLP:conf/naacl/MihalceaS05} created a one-liner dataset with humorous examples extracted from webpages with humor theme and non-humorous examples from Reuters titles, British National Corpus (BNC) sentences, and English Proverbs. \citet{DBLP:conf/emnlp/YangLDH15} scraped puns from the Pun of the Day website\footnote{\url{http://www.punoftheday.com/}} and negative examples from various news websites. There is also work on the curation of non-English humor datasets~\cite{DBLP:conf/emnlp/ZhangZLLX19, DBLP:conf/acl/BlinovBB19}. \citet{DBLP:conf/emnlp/HasanRZZTMH19} developed UR-FUNNY, a multimodal humor dataset that involves text, audio and video information extracted from TED talks.

\paragraph{Humor Recognition}
Most of the existing work on humor recognition in text focuses on one-liners, one type of jokes that delivers the laughter in a single line. The methodologies typically fall into two categories: feature engineering and deep learning. \citet{DBLP:conf/naacl/MihalceaS05} designed three human-centric features (alliteration, antonymy and synonym) for recognizing humor in the curated one-liner dataset. \citet{DBLP:conf/cicling/MihalceaSP10} approached the problem by calculating the semantic relatedness between the set-up and the punchline (they evaluated 150 one-liners by manually splitting them into ``set-up'' and ``punchline''). \citet{DBLP:conf/kdd/ShahafHM15} investigated funny captions for cartoons and proposed several features including perplexity to distinguish between funny and less funny captions. \citet{DBLP:conf/emnlp/MoralesZ17} proposed a probabilistic model and leveraged background text sources (such as Wikipedia) to identify humorous Yelp reviews. \citet{DBLP:conf/acl/LiuZS18} proposed to model sentiment association between elementary discourse units and designed features based on discourse relations. \citet{DBLP:conf/coling/CattleM18} explored the usage of word associations as a semantic relatedness feature in a binary humor classification task. With neural networks being popular in recent years, some deep learning structures have been developed for the recognition of humor in text. \citet{DBLP:journals/corr/ChenL17a} and \citet{DBLP:conf/naacl/ChenS18} adopted convolutional neural networks, while \citet{DBLP:conf/emnlp/WellerS19} used a Transformer architecture to do the classification task. \citet{DBLP:journals/complexity/FanLYDSCZ20} incorporated extra phonetic and semantic (ambiguity) information into the deep learning framework. In addition to these methodological papers, there are also some tasks dedicated to computational humor in recent years. SemEval 2020 Task 7~\cite{DBLP:conf/semeval/HossainKGK20} aims at assessing humor in edited news headlines. SemEval 2021 Task 7~\cite{meaney2021hahackathon} involves predicting the humor rating of the given text, and if the rating is controversial or not. In this task, \citet{xie2021semeval} adopted the DeBERTa architecture~\cite{DBLP:journals/corr/abs-2006-03654} with disentangled attention mechanism to predict the humor labels.

Although the work of \citet{DBLP:conf/cicling/MihalceaSP10} is the closest to ours, we are the first to bridge the incongruity theory of humor and large-scale pre-trained language models. Other work~\cite{DBLP:conf/naacl/BerteroF16} has attempted to predict punchlines in conversations extracted from TV series, but their subject of investigation should be inherently different from ours---punchlines in conversations largely depend on the preceding utterances, while jokes are much more succinct and self-contained.


\section{Humor Theories}
The attempts to explain humor date back to the age of ancient Greece, where philosophers like Plato and Aristotle regarded the enjoyment of comedy as a form of scorn, and held critical opinions towards laughter. These philosophical comments on humor were summarized as the \emph{superiority theory}, which states that laughter expresses a feeling of superiority over other people's misfortunes or shortcomings. Starting from the 18\textsuperscript{th} century, two other humor theories began to challenge the dominance of the superiority theory: the \emph{relief theory} and the \emph{incongruity theory}. The relief theory argues that laughter serves to facilitate the relief of pressure for the nervous system~\cite{sep-humor}. This explains why laughter is caused when people recognize taboo subjects---one typical example is the wide usage of sexual terms in jokes. The incongruity theory, supported by~\citet{kant1790critique}, \citet{schopenhauer1883world}, and many later philosophers and psychologists, states that laughter comes from the perception of something incongruous that violates the expectations. This view of humor fits well the types of jokes commonly found in stand-up comedies, where the set-up establishes an expectation, and then the punchline violates it. As an expansion of the incongruity theory, \citet{raskin1979semantic} proposed the Semantic Script-based Theory of Humor (SSTH) by applying the semantic script theory. It posits that, in order to produce verbal humor, two requirements should be fulfilled: (1) The text is compatible with two different scripts; (2) The two scripts with which the text is compatible are opposite.

\section{Methodology}
\label{sec:method}
The incongruity theory attributes humor to the violation of expectation. This means the punchline delivers the incongruity that turns over the expectation established by the set-up, making it possible to interpret the set-up in a completely different way. With neural networks blooming in recent years, pre-trained language models make it possible to study such relationship between the set-up and the punchline based on large-scale corpora. Given the set-up, language models are capable of writing expected continuations, enabling us to measure the degree of incongruity, by comparing the actual punchline with what the language model is likely to generate.

In this paper, we leverage the GPT-2 language model~\cite{radford2019language}, a Transformer-based architecture trained on the WebText dataset. We chose GPT-2 because: (1) GPT-2 is already pre-trained on massive data and publicly available online, which spares us the training process; (2) it is domain independent, thus suitable for modeling various styles of English text. Our goal is to model the set-up and the punchline as a whole piece of text using GPT-2, and analyze the probability of generating the punchline given the set-up. In the following text, we denote the set-up as $x$, and the punchline as $y$. Basically, we are interested in two quantities regarding the probability distribution $p(y|x)$: uncertainty and surprisal, which are elaborated in the next two sections.
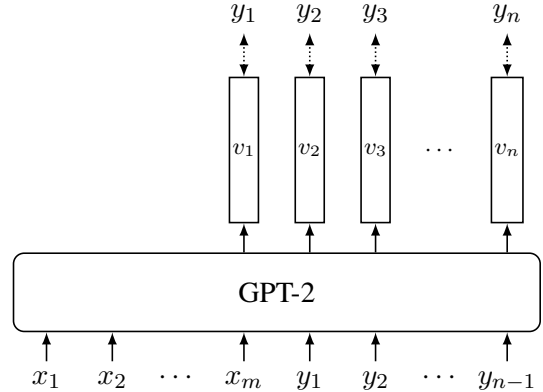
\begin{figure}[t]
    \centering
    \begin{tikzpicture}
        \node[rectangle,rounded corners,semithick,draw,align=center,inner sep=1em,minimum width=0.9\columnwidth] (gpt2) {GPT-2};
        \node[below left=1em and 1.125em of gpt2.south,anchor=north] (xm) {$x_m$};
        \node[below left=1em and 3.375em of gpt2.south,anchor=north] (xdot) {$\cdots$};
        \node[below left=1em and 5.625em of gpt2.south,anchor=north] (x2) {$x_2$};
        \node[below left=1em and 7.875em of gpt2.south,anchor=north] (x1) {$x_1$};
        \node[below right=1em and 1.125em of gpt2.south,anchor=north] (y1) {$y_1$};
        \node[below right=1em and 3.375em of gpt2.south,anchor=north] (y2) {$y_2$};
        \node[below right=1em and 5.625em of gpt2.south,anchor=north] (ydot) {$\cdots$};
        \node[below right=1em and 7.875em of gpt2.south,anchor=north] (yn) {$y_{n-1}$};
        \draw[-latex,semithick] (xm.north) -- ++ (0,1em);
        \draw[-latex,semithick] (x2.north) -- ++ (0,1em);
        \draw[-latex,semithick] (x1.north) -- ++ (0,1em);
        \draw[-latex,semithick] (yn.north) -- ++ (0,1em);
        \draw[-latex,semithick] (y2.north) -- ++ (0,1em);
        \draw[-latex,semithick] (y1.north) -- ++ (0,1em);
        \node[rectangle,semithick,draw,align=center,minimum height=5em,inner sep=0.1em] (v1) [above left=1em and 1.125em of gpt2.north,anchor=south] {\small $v_1$};
        \node[rectangle,semithick,draw,align=center,minimum height=5em,inner sep=0.1em] (v2) [above right=1em and 1.125em of gpt2.north,anchor=south] {\small $v_2$};
        \node[rectangle,semithick,draw,align=center,minimum height=5em,inner sep=0.1em] (v3) [above right=1em and 3.375em of gpt2.north,anchor=south] {\small $v_3$};
        \node[rectangle,align=center,minimum height=5em,inner sep=0.1em] (vdot) [above right=1em and 5.625em of gpt2.north,anchor=south] {\small $\cdots$};
        \node[rectangle,semithick,draw,align=center,minimum height=5em,inner sep=0.1em] (vn) [above right=1em and 7.875em of gpt2.north,anchor=south] {\small $v_n$};
        \draw[-latex,semithick] ([xshift=-1.125em]gpt2.north) -- ++ (0,1em);
        \draw[-latex,semithick] ([xshift=1.125em]gpt2.north) -- ++ (0,1em);
        \draw[-latex,semithick] ([xshift=3.375em]gpt2.north) -- ++ (0,1em);
        \draw[-latex,semithick] ([xshift=7.875em]gpt2.north) -- ++ (0,1em);
        \node[above=1.5em of v1] (y1a) {$y_1$};
        \node[above=1.5em of v2] (y2a) {$y_2$};
        \node[above=1.5em of v3] (y3a) {$y_3$};
        \node[above=1.5em of vn] (yna) {$y_n$};
        \draw[latex-latex,densely dotted,semithick] (v1) -- (y1a);
        \draw[latex-latex,densely dotted,semithick] (v2) -- (y2a);
        \draw[latex-latex,densely dotted,semithick] (v3) -- (y3a);
        \draw[latex-latex,densely dotted,semithick] (vn) -- (yna);
    \end{tikzpicture}
    \caption{The set-up $x$ and the punchline $y$ are concatenated and fed into GPT-2 for predicting the next token. $v_i$'s are probability distributions on the vocabulary.}
    \label{fig:gpt2}
\end{figure}

\subsection{Uncertainty}
The first question we are interested in is: given the set-up, how uncertain it is for the language model to continue? This question is related to SSTH, which states that, for a piece of text to be humorous, it should be compatible with two different scripts. To put it under the framework of set-up and punchline, this means the set-up could have multiple ways of interpretation, according to the following punchline. Thus, one would expect a higher uncertainty value when the language model tries to continue the set-up and generate the punchline.

We propose to calculate the averaged entropy of the probability distributions at all token positions of the punchline, to represent the degree of uncertainty. As shown in Figure~\ref{fig:gpt2}, the set-up $x$ and the punchline $y$ are concatenated and then fed into GPT-2 to predict the next token. While predicting the tokens of $y$, GPT-2 produces a probability distribution $v_i$ over the vocabulary. The averaged entropy is then defined as
\begin{equation}
    U(x,y) = -\frac{1}{|y|}\sum_{i=1}^n\sum_{w\in V}v_i^w\log v_i^w,
\end{equation}
where $V$ is the vocabulary.

\subsection{Surprisal}
The second question we would like to address is: how surprising it is when the language model actually generates the punchline? As the incongruity theory states, laughter is caused when something incongruous is observed and it violates the previously established expectation. Therefore, we expect the probability of the language model generating the actual punchline to be relatively low, which indicates the surprisal value should be high. Formally, the surprisal is defined as
\begin{equation}
\begin{split}
    S(x,y) &= -\frac{1}{|y|}\log p(y|x) \\
    &= -\frac{1}{|y|}\sum_{i=1}^n \log v_i^{y_i}.
\end{split}
\end{equation}

\section{Experiments}
We evaluated and compared the proposed features with several baselines by conducting experiments in two settings: predicting using individual features, and combining the features with a content-based text classifier.

\subsection{Baselines}
Similar to our approach of analyzing the relationship between the set-up and the punchline, \citet{DBLP:conf/cicling/MihalceaSP10} proposed to calculate the semantic relatedness between the set-up and the punchline. The intuition is that the punchline (which delivers the surprise) will have a minimum relatedness to the set-up. For our experiments, we chose two relatedness metrics that perform the best in their paper as our baselines, plus another similarity metric based on shortest paths in WordNet~\cite{DBLP:journals/cacm/Miller95}:
\begin{itemize}
    \item \textbf{Leacock \& Chodorow similarity} \cite{leacock1998combining}, defined as
    \begin{equation}
        \text{Sim}_{\textit{lch}} = -\log\frac{\textit{length}}{2*D},
    \end{equation}
    where \textit{length} is the length of the shortest path between two concepts using node-counting, and $D$ is the maximum depth of WordNet.

    \item \textbf{Wu \& Palmer similarity} \cite{10.3115/981732.981751} calculates similarity by considering the depths of the two synsets in WordNet, along with the depth of their \textit{LCS} (Least Common Subsumer), which is defined as
    \begin{equation}
        \text{Sim}_{\textit{wup}} = \frac{2*\text{depth}(\textit{LCS})}{\text{depth}(C_1) + \text{depth}(C_2)},
    \end{equation}
    where $C_{1}$ and $C_{2}$ denote synset 1 and synset 2 respectively.

    \item \textbf{Path similarity} \cite{DBLP:journals/tsmc/RadaMBB89} is also based on the length of the shortest path between two concepts in WordNet, which is defined as
    \begin{equation}
        \text{Sim}_{\textit{path}} = \frac{1}{1+\textit{length}}.
    \end{equation}
\end{itemize}
In addition to the metrics mentioned above, we also consider the following two baselines related to the phonetic and semantic styles of the input text:
\begin{itemize}
    \item \textbf{Alliteration}. The alliteration value is computed as the total number of alliteration chains and rhyme chains found in the input text \cite{DBLP:conf/naacl/MihalceaS05}.
    \item \textbf{Ambiguity}. Semantic ambiguity is found to be a crucial part of humor~\cite{DBLP:conf/acl/MillerG15}. We follow the work of~\citet{DBLP:conf/acl/LiuZS18} to compute the ambiguity value:
    \begin{equation}
        \log\prod_{w\in s}\text{num\_of\_senses}(w),
    \end{equation}
    where $w$ is a word in the input text $s$.
\end{itemize}

\subsection{Dataset}
We took the dataset from SemEval 2021 Task 7.\footnote{\url{https://semeval.github.io/SemEval2021/}} The released training set contains 8,000 manually labeled examples in total, with 4,932 being positive, and 3,068 negative. To adapt the dataset for our purpose, we only considered positive examples with exactly two sentences, and negative examples with at least two sentences. For positive examples (jokes), the first sentence was treated as the set-up and the second the punchline. For negative examples (non-jokes), consecutive two sentences were treated as the set-up and the punchline, respectively.\footnote{We refer to them as set-up and punchline for the sake of convenience, but since they are not jokes, the two sentences are not real set-up and punchline.} After splitting, we cleaned the data with the following rules:
(1) We restricted the length of set-ups and punchlines to be under 20 (by counting the number of tokens);
(2) We only kept punchlines whose percentage of alphabetical letters is greater than or equal to 75\%;
(3) We discarded punchlines that do not begin with an alphabetical letter.
 As a result, we obtained 3,341 examples in total, consisting of 1,815 jokes and 1,526 non-jokes. To further balance the data, we randomly selected 1,526 jokes, and thus the final dataset contains 3,052 labeled examples in total. For the following experiments, we used 10-fold cross validation, and the averaged scores are reported.

\subsection{Predicting Using Individual Features}
\begin{table}[t]
    \small
    \centering
    \begin{tabular}{lcccc}
        \toprule
         & P & R & F1 & Acc \\
        \midrule
        Random & 0.4973 & 0.4973 & 0.4958 & 0.4959 \\
        \midrule
        $\text{Sim}_\textit{lch}$ & 0.5291 & 0.5179 & 0.4680 & 0.5177 \\
        $\text{Sim}_\textit{wup}$ & 0.5289 & 0.5217 & 0.4919 & 0.5190 \\
        $\text{Sim}_\textit{path}$ & 0.5435 & 0.5298 & 0.4903 & 0.5291 \\
        Alliteration & 0.5353 & 0.5349 & 0.5343 & 0.5354 \\
        Ambiguity & 0.5461 & 0.5365 & 0.5127 & 0.5337 \\
        \midrule
        Uncertainty & 0.5840 & 0.5738 & 0.5593 & 0.5741 \\
        Surprisal & 0.5617 & 0.5565 & 0.5455 & 0.5570 \\
        \midrule
        U+S & \textbf{0.5953} & \textbf{0.5834} & \textbf{0.5695} & \textbf{0.5832} \\
        \bottomrule
    \end{tabular}
    \caption{Performance of individual features. Last row (U+S) is the combination of uncertainty and surprisal. P: Precision, R: Recall, F1: F1-score, Acc: Accuracy. P, R, and F1 are macro-averaged, and the scores are reported on 10-fold cross validation.}
    \label{tab:individual_features}
\end{table}
To test the effectiveness of our features in distinguishing jokes from non-jokes, we built an SVM classifier (parameters can be found in Appendix~\ref{sec:model_params}) for each individual feature (uncertainty and surprisal, plus the baselines). The resulted scores are reported in Table~\ref{tab:individual_features}. Compared with the baselines, both of our features (uncertainty and surprisal) achieved higher scores for all the four metrics. In addition, we also tested the performance of uncertainty combined with surprisal (last row of the table), and the resulting classifier shows a further increase in the performance. This suggests that, by jointly considering uncertainty and surprisal of the set-up and the punchline, we are better at recognizing jokes.

\subsection{Boosting a Content-Based Classifier}
\begin{table}[t]
    \small
    \centering
    \begin{tabular}{lcccc}
        \toprule
         & P & R & F1 & Acc \\
        \midrule
        GloVe & 0.8233 & 0.8232 & 0.8229 & 0.8234 \\
        \midrule
        GloVe+$\text{Sim}_\textit{lch}$ & 0.8255 & 0.8251 & 0.8247 & 0.8250 \\
        GloVe+$\text{Sim}_\textit{wup}$ & 0.8264 & 0.8260 & 0.8254 & 0.8257 \\
        GloVe+$\text{Sim}_\textit{path}$ & 0.8252 & 0.8244 & 0.8239 & 0.8244 \\
        GloVe+Alliter. & 0.8299 & 0.8292 & 0.8291 & 0.8297 \\
        GloVe+Amb. & 0.8211 & 0.8203 & 0.8198 & 0.8201 \\
        \midrule
        GloVe+U & 0.8355 & 0.8359 & 0.8353 & 0.8359 \\
        GloVe+S & 0.8331 & 0.8326 & 0.8321 & 0.8326 \\
        \midrule
        GloVe+U+S & \textbf{0.8368} & \textbf{0.8368} & \textbf{0.8363} & \textbf{0.8365} \\
        \bottomrule
    \end{tabular}
    \caption{Performance of the features when combined with a content-based classifier. U denotes uncertainty and S denotes surprisal. P: Precision, R: Recall, F1: F1-score, Acc: Accuracy. P, R, and F1 are macro-averaged, and the scores are reported on 10-fold cross validation.}
    \label{tab:combined_features}
\end{table}
Now that we have shown the advantage of our features when used individually in prediction, we would like to validate their effectiveness when combined with the commonly used word embeddings. Thus, we evaluated our features as well as the baselines under the framework of a content-based classifier. The idea is to see if the features could further boost the performance of existing text classifiers. To create a starting point, we encoded each set-up and punchline into vector representations by aggregating the GloVe~\cite{DBLP:conf/emnlp/PenningtonSM14} embeddings of the tokens (sum up and then normalize by the length). We used the GloVe embeddings with dimension 50, and then concatenated the set-up vector and the punchline vector, to represent the whole piece of text as a vector of dimension 100. For each of the features (uncertainty and surprisal, plus the baselines), we appended it to the GloVe vector, and built an SVM classifier to do the prediction. Scores are reported in Table~\ref{tab:combined_features}. As we can see, compared with the baselines, our features produce larger increases in the performance of the content-based classifier, and similar to what we have observed in Table~\ref{tab:individual_features}, jointly considering uncertainty and surprisal gives further increase in the performance.

\section{Visualizing Uncertainty and Surprisal}
\begin{figure}[t]
    \centering
    \includegraphics[width=0.45\columnwidth]{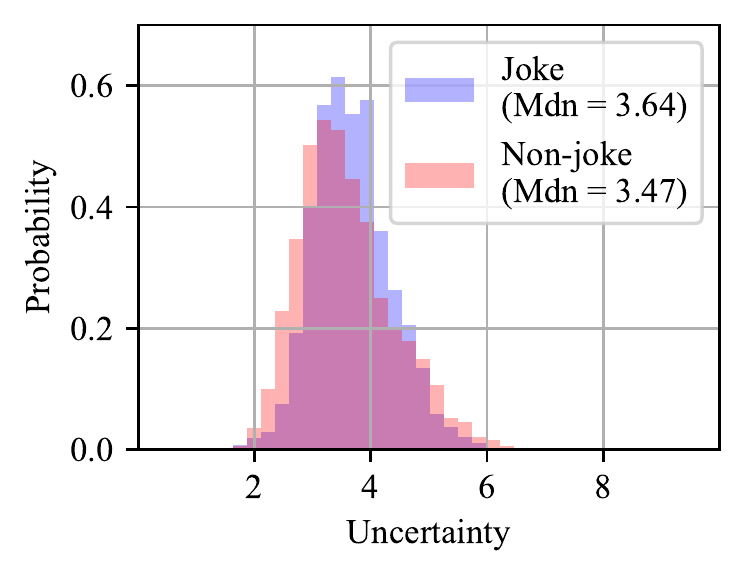}\quad
    \includegraphics[width=0.45\columnwidth]{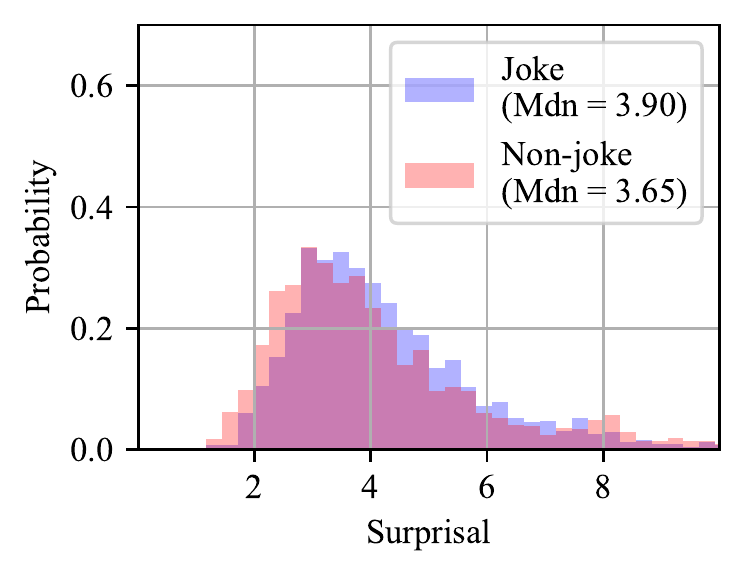}
    \caption{Histograms of uncertainty (left) and surprisal (right), plotted separately for jokes and non-jokes. Mdn stands for Median.}
    \label{fig:histograms}
\end{figure}
To get a straightforward vision of the uncertainty and surprisal values for jokes versus non-jokes, we plot their histograms in Figure~\ref{fig:histograms} (for all 3,052 labeled examples). It can be observed that, for both uncertainty and surprisal, jokes tend to have higher values than non-jokes, which is consistent with our expectations in Section~4.

\section{Conclusion}
This paper makes an attempt in establishing a connection between the humor theories and the nowadays popular pre-trained language models. We proposed two features according to the incongruity theory of humor: uncertainty and surprisal. We conducted experiments on a humor dataset, and the results suggest that our approach has an advantage in humor recognition over the baselines. The proposed features can also provide insight for the task of two-line joke generation---when designing the text generation algorithm, one could exert extra constraints so that the set-up is chosen to be compatible with multiple possible interpretations, and the punchline should be surprising in a way that violates the most obvious interpretation. We hope our work could inspire future research in the community of computational humor.

\bibliography{acl2021}
\bibliographystyle{acl_natbib}

\appendix
\section{Model Parameters}
\label{sec:model_params}
For the SVM classifier, we set the regularization parameter $C=1.0$, and used the RBF kernel with the kernel coefficient $\gamma=1/n_\text{features}$. All models were trained and evaluated on a machine with Intel Core i7-6700K CPU, Nvidia GeForce GTX 1080 GPU, and 16GB RAM. The running time of each method is listed in Table~\ref{tab:running_time_ind} and Table~\ref{tab:running_time_comb}.
\begin{table}[t]
    \centering
    \begin{tabular}{lr}
        \toprule
        & \textbf{Running Time} \\
        \midrule
        $\text{Sim}_\textit{lch}$ & 1.76 sec \\
        $\text{Sim}_\textit{wup}$ & 1.71 sec \\
        $\text{Sim}_\textit{path}$ & 1.71 sec \\
        Alliteration & 1.70 sec \\
        Ambiguity & 2.94 sec \\
        \midrule
        Uncertainty & 2.12 sec \\
        Surprisal & 2.49 sec \\
        \midrule
        Uncertainty + Surprisal & 2.26 sec \\
        \bottomrule
    \end{tabular}
    \caption{Running time of the SVM classifiers trained on individual features.}
    \label{tab:running_time_ind}
\end{table}
\begin{table}[t]
    \centering
    \begin{tabular}{lr}
        \toprule
        & \textbf{Running Time} \\
        \midrule
        GloVe & 7.54 sec \\
        \midrule
        GloVe + $\text{Sim}_\textit{lch}$ & 14.85 sec \\
        GloVe + $\text{Sim}_\textit{wup}$ & 15.90 sec \\
        GloVe + $\text{Sim}_\textit{path}$ & 13.76 sec \\
        GloVe + Alliteration & 15.41 sec \\
        GloVe + Ambiguity & 14.28 sec \\
        \midrule
        GloVe + Uncertainty & 14.70 sec \\
        GloVe + Surprisal & 13.84 sec \\
        \midrule
        GloVe + U + S & 19.27 sec \\
        \bottomrule
    \end{tabular}
    \caption{Running time of the content-based SVM classifiers combined with individual features. U denotes uncertainty and S denotes surprisal.}
    \label{tab:running_time_comb}
\end{table}

\end{document}